\documentclass[conference]{IEEEtran}
\IEEEoverridecommandlockouts
\usepackage{cite}
\usepackage{amsmath,amssymb,amsfonts}
\usepackage{algorithmic}
\usepackage{graphicx}
\usepackage{textcomp}
\usepackage{xcolor}
\usepackage{multirow}
\usepackage{caption}
\usepackage{subfig}

\def\BibTeX{{\rm B\kern-.05em{\sc i\kern-.025em b}\kern-.08em
    T\kern-.1667em\lower.7ex\hbox{E}\kern-.125emX}}
\begin{document}

\title{FHDR: HDR Image Reconstruction from a Single LDR Image using Feedback Network\\
}

\author{\IEEEauthorblockN{Zeeshan Khan}
\IEEEauthorblockA{
\textit{Indian Institute of Technology}\\
Gandhinagar, India \\
zeeshank606@gmail.com}
\and
\IEEEauthorblockN{Mukul Khanna}
\IEEEauthorblockA{
\textit{Indian Institute of Technology}\\
Gandhinagar, India \\
mukul18khanna@gmail.com}
\and
\IEEEauthorblockN{Shanmuganathan Raman}
\textit{Indian Institute of Technology}\\
Gandhinagar, India \\
shanmuga@iitgn.ac.in}

\maketitle

\begin{abstract}
High dynamic range (HDR) image generation from a single exposure low dynamic range (LDR) image has been made possible due to the recent advances in Deep Learning. Various feed-forward Convolutional Neural Networks (CNNs) have been proposed for learning LDR to HDR representations. To better utilize the power of CNNs, we exploit the idea of feedback, where the initial low level features are guided by the high level features using a hidden state of a Recurrent Neural Network. 
 Unlike a single forward pass in a conventional feed-forward network, the reconstruction from LDR to HDR in a feedback network is learned over multiple iterations. This enables us to create a coarse-to-fine representation, leading to an improved reconstruction at every iteration.
Various advantages over standard feed-forward networks include early reconstruction ability and better reconstruction quality with fewer network parameters. We design a dense feedback block and propose an end-to-end feedback network- FHDR for HDR image generation from a single exposure LDR image. Qualitative and quantitative evaluations show the superiority of our approach over the state-of-the-art methods.

\end{abstract}


\section{Introduction}
Common digital cameras can not capture the wide range of light intensity levels in a natural scene.
This can lead to a loss of pixel information in under-exposed and over-exposed regions of an image, resulting in a low dynamic range (LDR) image. To recover the lost information and represent the wide range of illuminance in an image, high dynamic range (HDR) images need to be generated. There has been active research going on in the area of deep learning for HDR imaging. The advances in deep learning for image processing tasks have paved way for various approaches for HDR image reconstruction using feed-forward convolutional neural network (CNN) architectures \cite{eilertsen2017hdr}\cite{marnerides2018expandnet}\cite{endo2017deep}\cite{yang2018image}\cite{lee2018deep}. The above methods specifically transform a single exposure LDR image into an HDR image. 
HDRCNN \cite{eilertsen2017hdr} proposed a deep autoencoder for HDR image reconstruction which uses a weighted mask to recover only the over-exposed regions of an LDR image. The authors of DRTMO \cite{endo2017deep} designed a framework with two networks for generating up-exposure and down-exposure LDR images, which are merged to form an HDR image. The network is not end-to-end trainable and uses a large number of parameters. Unlike the others, the proposed FHDR model provides an end-to-end trainable solution and is able to comprehensively learn the LDR-HDR mapping while outperforming the existing methods.

Deeper networks are known to learn more complex non-linear relationships like the LDR to HDR mapping. The caveat with deeper networks is that they consume a lot of computational resources and tend to over-fit the training data. To overcome this problem, we 
exploit the power of feedback mechanisms, inspired by \cite{zamir2017feedback}, for the task of HDR image reconstruction. A feedback block is an RNN whose output is fed back to guide its input via a hidden state.
A feedback network can run for many iterations for a single training example. Considering the number of iterative operations on the shared network parameters, a feedback network is virtually deeper than the corresponding feed-forward network with the same physical depth. Here, virtual depth = physical depth $\times$ number of iterations.

For improving the reconstruction at every iteration, the loss is calculated for the output of each iteration.
By doing so, the network is forced to create a coarse-to-fine representation, being able to reconstruct HDR content right from the first iteration and improve with every subsequent iteration.
We propose a global feedback block which consists of smaller local feedback blocks of densely connected layers, inspired by the Dense-Net architecture \cite{huang2017densely}. Dense connections allow the network to reuse features. This helps in learning robust image representations, even with lesser network parameters. 

The performance of our framework is evaluated on the standard City Scene dataset \cite{zhang2017learning} and another dataset prepared from the list of HDR image datasets suggested in \cite{eilertsen2017hdr}. 
Qualitative and quantitative assessments of the network suggest that even with fewer network parameters, the proposed FHDR model outperforms the state-of-the-art methods.


\section{HDR reconstruction framework}
\subsection{Feedback system}
Feedback systems are adopted to influence the input based on the generated output, unlike the conventional feed-forward networks, where information flow is unidirectional and is not directly influenced by the generated output.
The authors in \cite{zamir2017feedback} exploited the feedback mechanism using an RNN, where the output of the feedback block travels back to guide its input via a hidden state. Their architecture uses ConvLSTM cells as the basic RNN units and is designed for the task of image classification. Even with lesser network parameters as compared to other feed-forward networks, such networks are able to learn better representations.
Recently, authors of \cite{li2019srfbn} designed a feedback network specifically for the task of image super-resolution which achieved state-of-the-art performance.
\par 
Inspired by the success of feedback networks, we designed a feedback network for learning the LDR to HDR mapping that has been explained in detail in the following sections.
 
\subsection{Model architecture}
\begin{figure}[htbp]
\centering
 \includegraphics[scale=0.3]{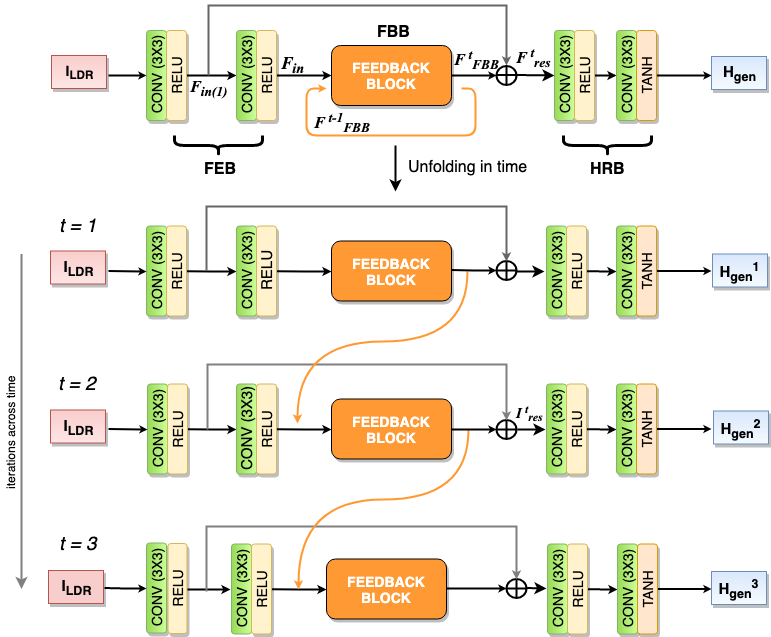}
\caption{FHDR Architecture}
\label{fig1}
\end{figure}

Our architecture consists of three blocks similar to \cite{li2019srfbn}, as shown in Fig. \ref{fig1}. The first block is the Feature Extraction block \textbf{(FEB)}, followed by the Feedback block \textbf{(FBB)} and an HDR reconstruction block \textbf{(HRB)}. Inspired by \cite{zhang2018residual}, we use a global residual skip connection for bypassing low level LDR features at every iteration to guide the HDR reconstruction block in the final layers.
For every training example, the network runs for $n$ iterations. Here each iteration from $t=1$ to $t=n$ is a forward pass in time in an unfolded RNN.
The FEB is responsible for extracting the low-level feature information $F_{in}$ from the input LDR image $I_{LDR}$.

\begin{equation}
F_{in} = f_{FEB}(I_{LDR})\label{eq}.
\end{equation}

Here, $f_{FEB}$ represents the operations of the FEB.
To achieve the feedback mechanism, $F_{in}$ is fed to the FBB, combined with the output of the FBB from the previous iteration, using a global hidden state as below. 
\begin{equation}
F^t_{FBB} = f_{FBB}(F_{in}, F^{t-1}_{FBB}).\label{eq}
\end{equation}
Here, $F^t_{FBB}$ represents the output of the feedback block at iteration $t$. At $t=1$, when there is no feedback, the hidden state is initialised with the values of the extracted features $F_{in}$.

At every iteration, the low level LDR features from the first convolutional layer in FEB are added to the output of the FBB using a global residual skip connection as below.
\begin{equation}
F^t_{res} = F_{in(1)}+ F_{FBB}^t\label{eq}.
\end{equation}
Here, $F^t_{res}$ represents the global residual feature map learned at iteration $t$. $F_{in(1)}$ stands for the low level LDR features from the first convolutional layer in FEB. $F^t_{res}$ is passed to the HRB to generate an HDR image at every iteration as below.
\begin{equation}
H^t_{gen} = f_{HRB}(F^t_{res})\label{eq}
\end{equation}
Here, $f_{HRB}$ represents the operations of the HRB and $H^t_{gen}$ represents the HDR image generated at $t^{th}$ iteration.
For every LDR image, a forward pass can run for $n$ iterations, therefore generating $n$ HDR images. Each generated image is coupled with a loss, hence resulting in improved reconstruction at each iteration through back-propagation in time.

\subsection{Feedback block}

\begin{figure}[htbp]
\centering
\includegraphics[scale=0.34]{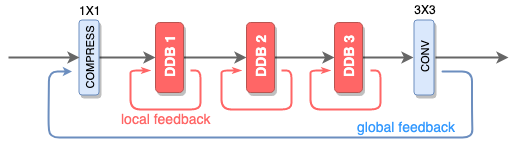}
\caption{Feedback block}
\label{fig2}
\end{figure}

\begin{figure}[htbp]
\centering
\includegraphics[scale=0.3]{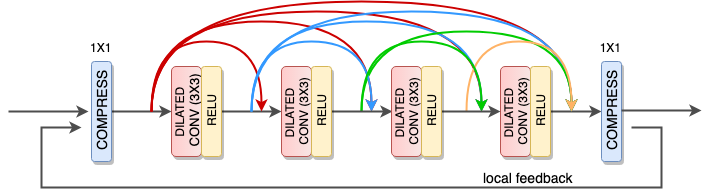}
\caption{Dilated Dense Block\\[-4ex]} 
\label{fig3}
\end{figure}

We have designed a novel feedback block for the task of learning LDR-to-HDR representations, as shown in Fig. \ref{fig2}. 
The basic unit of the feedback block is a Dilated Dense Block (DDB) shown in Fig. \ref{fig3}. It is a modification of the Dense block proposed in \cite{huang2017densely}. Dilated convolutions help in increasing the receptive field of the network \cite{yu2015multi}. 
A DDB helps in utilising all the hierarchical features from the input. 
Other than the two $1 \times 1$ convolutional layers for feature compression, each DDB houses four dilated $3 \times 3$ convolutional layers, each of which uses the information from all the previous layers using dense skip connections. This reuse of features due to the dense forward connections allows for reduced network parameters and improves the learning ability. Three of such DDBs come together to form the feedback block of the network.

We implement global and local feedback mechanisms described as follows.
\subsubsection{Global Feedback} The global feedback block, FBB is considered as an RNN with a global hidden state. High level features are transferred from the output of the feedback block at the $t-1^{th}$ iteration to its input at the $t^{th}$ iteration. The hidden state is concatenated with $FT_{in}$ and a $1 \times 1$ compression convolution layer is applied for high-level and low-level feature fusion as shown in Fig. \ref{fig2}. The fused features are passed to the dilated dense blocks, followed by a $3 \times 3$ convolution layer for further processing.

\subsubsection{Local Feedback} We argue that a feedback connection is always beneficial as it helps to guide the low-level features which are in some way blind to the higher level features. Hence, we have implemented local feedback connections over each DDB which aim to improve the features generated locally. These connections run parallel to the global feedback connections and increase the overall effectiveness of the network. Each DDB can be considered as an RNN similar to the global feedback block, transferring features from its output to its input via a local hidden state.\\[-3.5ex]

\subsection{Loss function}
Loss calculated directly on HDR images is misrepresented due to the dominance of high intensity values of images with a wide dynamic range. Therefore, we tonemap the generated and the ground truth HDR images to compress the wide intensity range before calculating the loss function value. We use the $\mu$-law for tonemapping, as suggested by \cite{kalantari2017deep}. The $\mu$-law is represented as below.
\begin{equation}
T(H^t_{gen}) =  \frac{log(1+\mu H^t_{gen})}{log(1+\mu)}\label{eq}
\end{equation}
Here, $T$ represents the tonemapping operation and $\mu$ defines the amount of compression, which is set to 5000 for the experiments. In addition to the L1 loss suggested by previous feedback networks, we use a perceptual loss \cite{johnson2016perceptual} for improving the visual quality of the generated image.
We calculate the L1 loss and the perceptual loss at every iteration $t$ and take an average over all the iterations. The average L1 loss $\mathcal{L}_{L1}$ is given below.
\begin{equation}
\mathcal{L}_{L1}= \frac{1}{n}\sum_{t=1}^{n} \big|\big| T(H^t_{gen}) - T(H_{gt}) \big|\big|
\end{equation}
Here, $H_{gt}$ represents the ground truth image. The average perceptual loss $\mathcal{L}_p$ can be represented as below.
\begin{equation}
\mathcal{L}_p = \frac{1}{n}\sum_{t=1}^{n}f_{VGG_{19}}\big(T(H^t_{gen}), T(H_{gt})\big)\label{eq}
\end{equation}
Here, $f_{VGG_{19}}$ represents the perceptual loss calculated between the tonemapped ground truth and generated images.
The final loss function $\mathcal{L}$ is given below.
\begin{equation}
\mathcal{L} =\mathcal{L}_p + \lambda \mathcal{L}_{L1}\label{eq}
\end{equation}
Here, $\lambda$ is set to 0.1 for all the experiments.
We have observed that applying only the L1 distance produces dark artifacts.
However, a combination of L1 loss and perceptual loss has resulted in improved visual quality of the generated image.

\subsection{Implementation details}

All the convolutional layers (conv2d) in the network have $3 \times 3$ kernels, followed by a ReLU activation, unless mentioned otherwise. There are 2 conv2d layers in the FEB, 20 ($6\times 3 +2$) conv2d layers in FBB, and 2 conv2d layers in the HRB. The size of the feature maps remains same throughout the network. The depth of the feature maps also remain same i.e. 64, except for in the DDBs.
A growth rate of 32 has been implemented for the DDBs, meaning that $3 \times 3$ conv2d layers of each dense block output a 32 channel feature map that gets concatenated with features from all the preceding layers. This accumulated feature map depth is brought down using a $1 \times 1$ conv2d layer at the end of the dense block.

We trained our network on Geforce RTX 2070 GPU with a batch-size of 16 for the City Scene dataset and 6 for the curated HDR dataset. Adam optimizer \cite{kingma2014adam} was adopted with momentum parameters $\beta_{1}$ = 0.5 and $\beta_{2}$ = 0.999. All the variants of the proposed model were trained for 200 epochs with an initial learning rate of $ 2 \times 10^{-4}$ for first 100 epochs, decayed linearly over the next 100 epochs. 



\begin{figure*}%
    \centering
    \subfloat{{\includegraphics[scale=0.118]{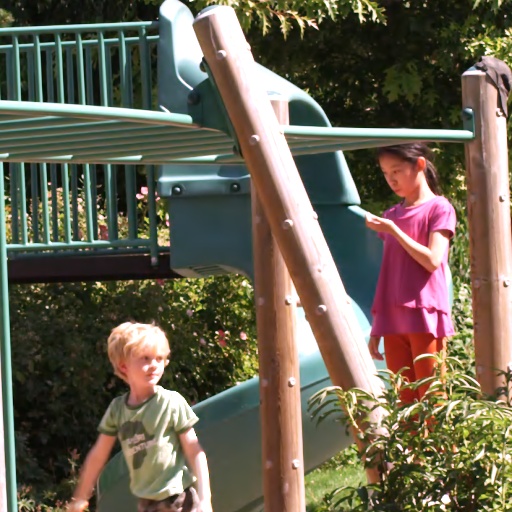} }}%
    \subfloat{{\includegraphics[scale=0.118]{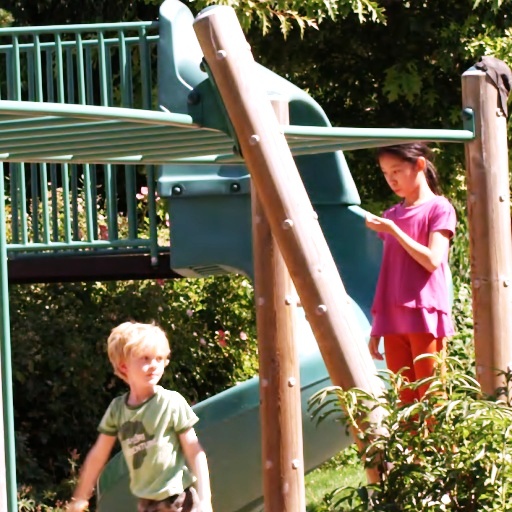} }}%
    \subfloat{{\includegraphics[scale=0.118]{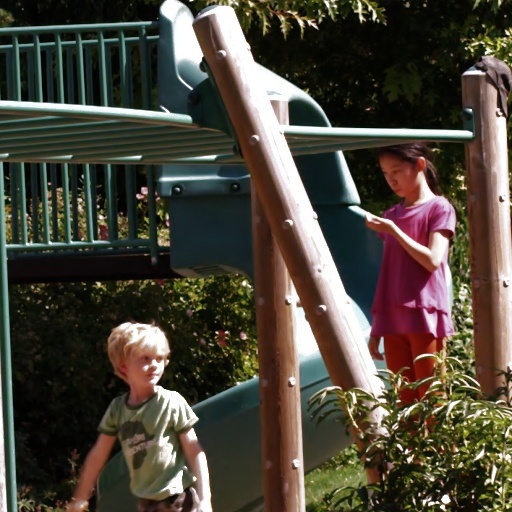} }}%
    \subfloat{{\includegraphics[scale=0.118]{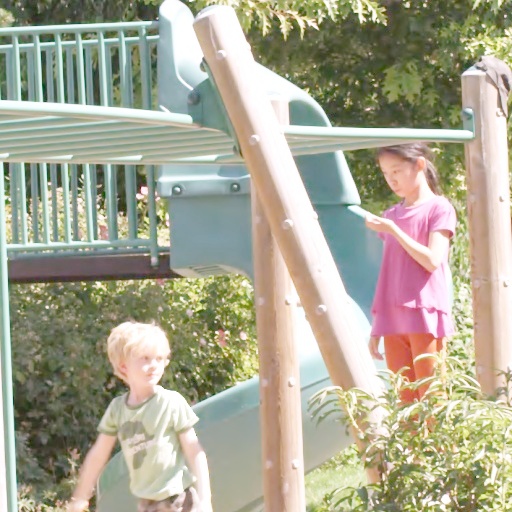} }}%
    \subfloat{{\includegraphics[scale=0.118]{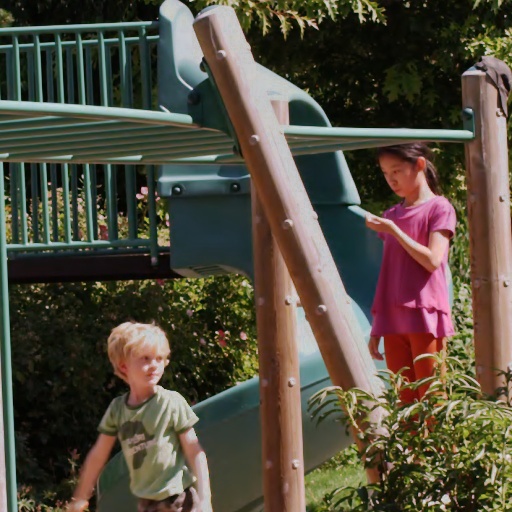} }}%
    \subfloat{{\includegraphics[scale=0.118]{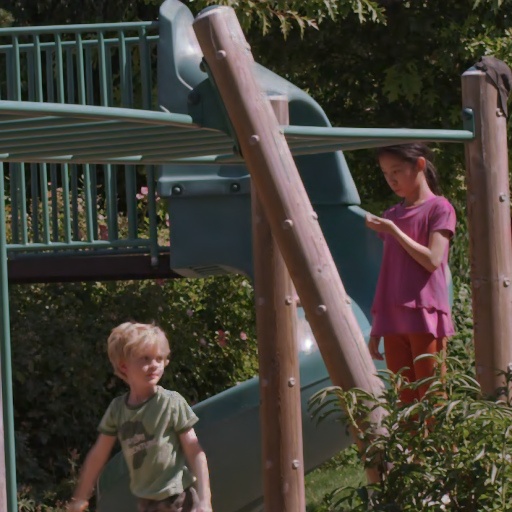} }}%
    \subfloat{{\includegraphics[scale=0.118]{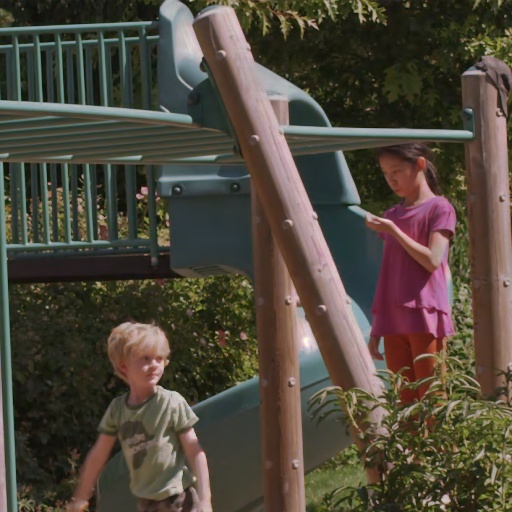} }}%
    \subfloat{{\includegraphics[scale=0.118]{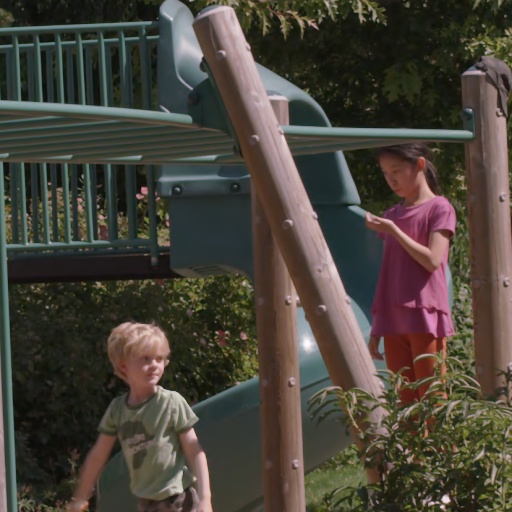} }}\\[-4.2ex]%
    \qquad
    \subfloat[LDR]{{\includegraphics[scale=0.118]{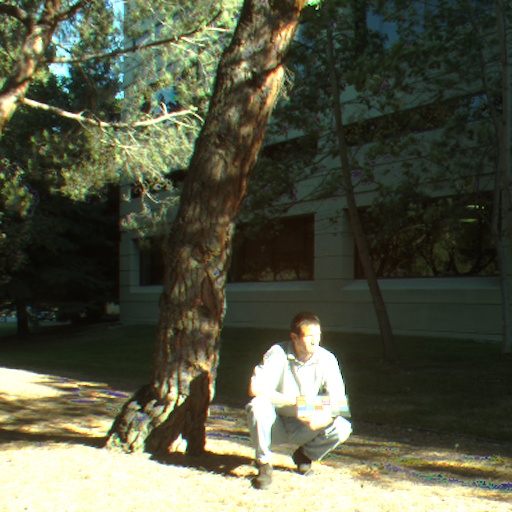} }}%
    \subfloat[AKY\cite{akyuz2007hdr}]{{\includegraphics[scale=0.118]{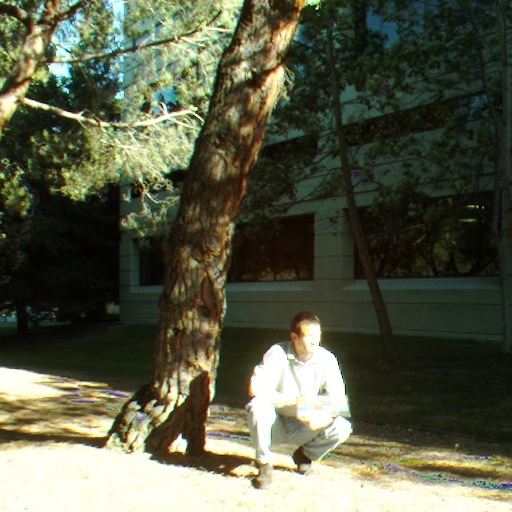} }}%
    \subfloat[KOV\cite{kovaleski2014high}]{{\includegraphics[scale=0.118]{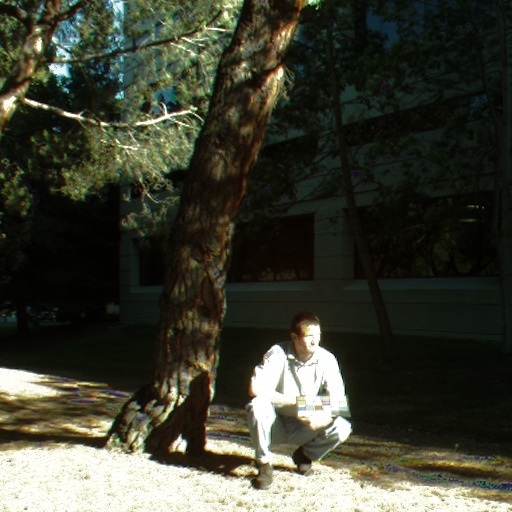} }}%
    \subfloat[DRTMO\cite{endo2017deep}]{{\includegraphics[scale=0.118]{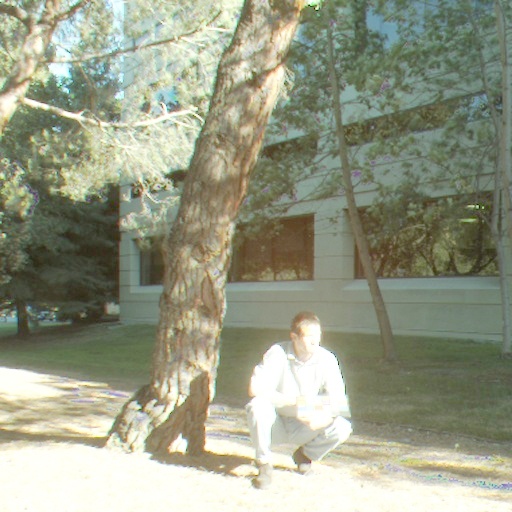} }}%
    \subfloat[HDRCNN\cite{eilertsen2017hdr}]{{\includegraphics[scale=0.118]{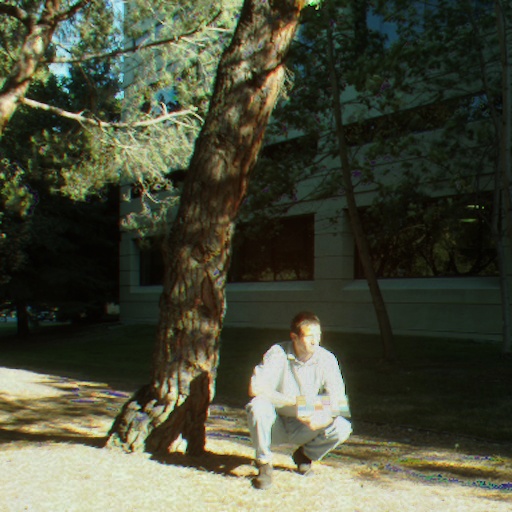} }}%
    \subfloat[FHDR/W]{{\includegraphics[scale=0.118]{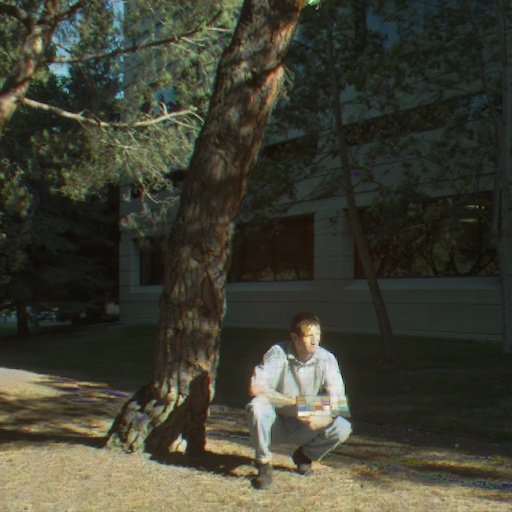} }}%
    \subfloat[FHDR]{{\includegraphics[scale=0.118]{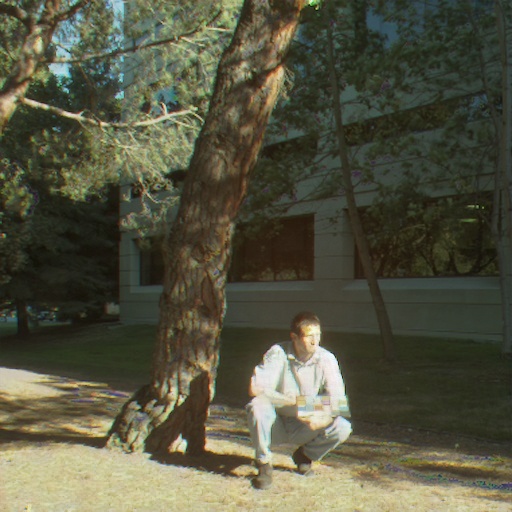} }}%
    \subfloat[GROUND TRUTH]{{\includegraphics[scale=0.118]{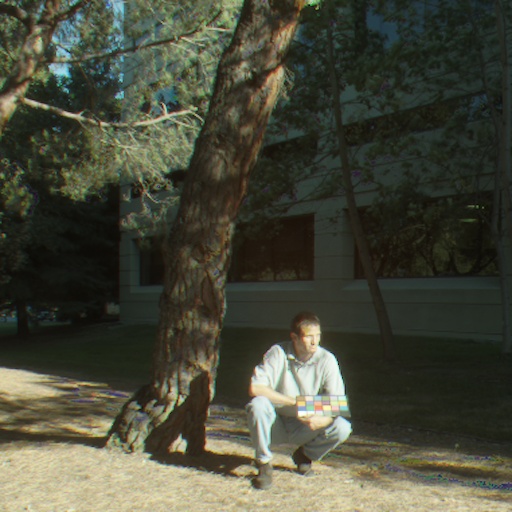} }}%
    \caption{Qualitative evaluation against five described methods on the curated HDR dataset. (HDR images have been tonemapped using Reinhard tonemapping algorithm \cite{Reinhard:2002:PTR:566654.566575} for displaying.)}
    \label{fig5}%
\end{figure*}

\section{Experiments}

\subsection{Dataset}

We have trained and evaluated the performance of our network on the standard City Scene dataset \cite{zhang2017learning} and another HDR dataset curated from various sources shared in \cite{eilertsen2017hdr}.
The City Scene dataset is a low resolution HDR dataset that contains pairs of LDR and ground truth HDR images of size $128 \times 64$. We trained the network on 39460 image pairs provided in the training set and evaluated it on 1672 randomly selected images from the testing set. The curated HDR dataset consists of 1010 HDR images of high resolution.
Training and testing pairs were created by producing LDR counterparts of the HDR images by exposure alteration and application of different camera curves from \cite{1211522}. Data augmentation was done using random cropping and resizing. The final training set consists of 11262 images of size $256 \times $256. The testing set consists of 500 images of size $512 \times $512 to evaluate the performance of the network on high resolution images.

\subsection{Evaluation metrics}
For the quantitative evaluation of the proposed method, we use the HDRVDP-2 Q-score metric, specifically designed for evaluating reconstruction of HDR images based on human perception \cite{narwaria2015hdr}. We also use the commonly adopted PSNR and SSIM image-quality metrics. The PSNR score is calculated in dB between the $\mu$-law tonemapped ground truth and generated images.

\subsection{Feedback mechanism analysis}
\begin{figure}[htbp]
\centering
\includegraphics[scale=0.35]{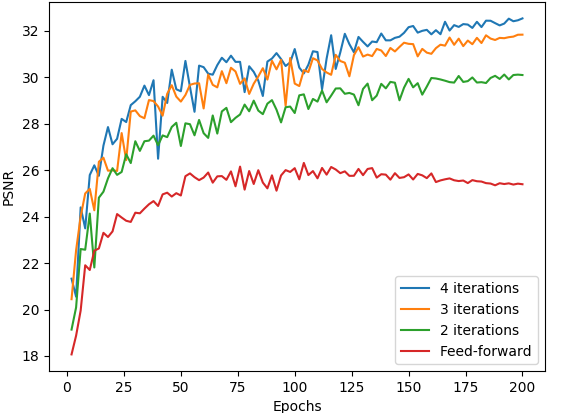}
\caption{Convergence analysis for four variants of FHDR, evaluated on the City Scene dataset.}
\label{fig4}
\end{figure}

To study the influence of feedback mechanisms, we compare the results of four variants of the network based on the number of feedback iterations performed - (i) FHDR/W (feed-forward / without feedback), (ii) FHDR-2 iterations, (iii) FHDR-3 iterations, and (iv) FHDR-4 iterations. To visualise the overall performance, we plot the PSNR score values against the number of epochs for the four variants, shown in Fig. \ref{fig4}. The significant impact of feedback connections can be easily observed. The PSNR score increases as the number of iterations increase. Also, early reconstruction can be seen in the network with feedback connections. Based on this, we decided to implement an iteration count of 4 for the proposed FHDR network.
\\[-4ex]

\section{RESULTS}
Qualitative and quantitative evaluations are performed against two non-learning based inverse tonemapping methods- (AKY) \cite{akyuz2007hdr} and (KOV) \cite{kovaleski2014high}, two deep learning methods- HDRCNN \cite{eilertsen2017hdr} and DRTMO \cite{endo2017deep} and the feed-forward counterpart of the proposed FHDR method. We use the HDR toolbox \cite{banterle2017advanced} for evaluating the non-learning based methods. 
The deep learning methods were trained and evaluated on the described datasets, as mentioned in earlier sections. 

DRHT \cite{yang2018image} is the state-of-the-art deep neural network for the image correction task. It reconstructs HDR content as an intermediate output and transforms it back into an LDR image. Due to unavailability of DRHT network implementation, results of their network on the curated dataset are not presented. Performance metrics of their LDR-to-HDR network, trained on the City Scene dataset has been sourced from their paper because of the same experiment setup in terms of the training and testing datasets.


\subsection{Quantitative evaluation}
A quantitative comparison of the above mentioned HDR reconstruction methods has been presented in Table 1. The proposed network outperforms the state-of-the-art methods in evaluations performed on both the datasets. Results of DRTMO \cite{endo2017deep} could not be calculated on the City Scene dataset because the network does not accept low resolution images. Even though the feed-forward counterpart of FHDR outperforms other methods, the proposed FHDR network (4-iterations) performs far better.

\subsection{Qualitative evaluation}
As can be seen in Fig. \ref{fig5}, non-learning based methods are unable to recover highly over-exposed regions. DRTMO brightens the whole image and is able to recover only the under-exposed, regions. HDRCNN, on the other hand, is designed to recover only the saturated regions and thus under performs in recovering information in under-exposed dark regions of the image. Unlike others, our FHDR/W pays equal attention to both under-exposed and over-exposed regions of the image. The proposed FHDR network with the feedback connections enhances the overall sharpness of the image and performs an even better reconstruction of the input. 


\begin{table}[htbp]
\centering
\begin{tabular}{|p{1.4cm}|c|c|c|c|c|c|}
\hline
\multirow{2}{*}{\textbf{Methods}} & \multicolumn{3}{|c|}{\textbf{City Scene Dataset}}&\multicolumn{3}{|c|}{\textbf{Curated HDR Dataset}} \\
\cline{2-7} 
& \textbf{PSNR}& \textbf{SSIM}& \textbf{Q-score} & \textbf{PSNR}& \textbf{SSIM}& \textbf{Q-score} \\
\hline
AKY\cite{akyuz2007hdr} & 15.35 & 0.44 & 35.40 & 9.58 & 0.20 & 33.47 \\
\hline
KOV\cite{kovaleski2014high} & 16.77 & 0.59 & 35.31 & 12.99 & 0.41 & 29.87 \\
\hline
HDRCNN\cite{eilertsen2017hdr} & 13.21 & 0.38 & 54.34 & 12.13 & 0.34 & 55.32 \\
\hline
DRTMO\cite{endo2017deep} & - & - & - & 11.4 & 0.28 & 58.85 \\
\hline
DRHT\cite{yang2018image} & - & 0.93 & 61.51 & - & - & - \\
\hline
FHDR/W & 25.39 & 0.89 & 63.21 & 16.94 & 0.74 & 65.27 \\
\hline
FHDR & \textbf{32.54} & \textbf{0.95} & \textbf{67.18} & \textbf{20.3} & \textbf{0.79} & \textbf{70.97} \\
\hline
\end{tabular}
\label{tab1}
\caption{Quantitative comparison against state-of-the-art methods.\\[-4.5ex]}
\end{table}
\section{Conclusion}
We propose a novel feedback network FHDR, to reconstruct an HDR image from a single exposure LDR image. The dense connections in the forward-pass enable feature-reuse, thus learning robust representations with minimum parameters. Local and global feedback connections enhance the learning ability, guiding the initial low level features from the high level features.
Iterative learning forces the network to create a coarse-to-fine representation which results in early reconstructions. 
Extensive experiments demonstrate that the FHDR network is successfully able to recover the underexposed and overexposed regions outperforming state-of-the-art methods.

\section{Acknowledgement}
This research was supported by the SERB Core Research Grant.

\bibliography{references} 
\bibliographystyle{ieeetr}
\end{document}